\documentclass[letterpaper]{article} 
\usepackage{aaai2027}  
\usepackage[hyphens]{url}  
\usepackage{graphicx} 
\usepackage{natbib}  
\usepackage{caption} 
\usepackage{algorithm}
\usepackage{algorithmic}

\usepackage{newfloat}
\usepackage{listings}
\DeclareCaptionStyle{ruled}{labelfont=normalfont,labelsep=colon,strut=off} 
\floatstyle{ruled}
\newfloat{listing}{tb}{lst}{}
\floatname{listing}{Listing}

\usepackage{booktabs}

\nocopyright 

\title{Unexplainability of Artificial Intelligence Judgments and Functional Implementation in Kant's Perspective}
\author{
    Jongwoo Seo
}
\affiliations{

    Busan, South Korea\\
}

\begin{document}

\maketitle

\begin{abstract}
Kant's \textit{Critique of Pure Reason}, a major contribution to the history of epistemology, proposes a table of categories to elucidate the structure of the a priori principles underlying human judgment. Artificial intelligence (AI) technology claims to simulate or replicate human judgment. To evaluate this claim, it is necessary to examine whether AI judgments exhibit the essential characteristics of human judgment. This paper investigates the unexplainability of AI judgments through the lens of Kant’s theory of judgment. Drawing on Kant’s four logical forms --- quantity, quality, relation, and modality --- this study identifies what may be called AI's uncertainty, a condition in which different forms of judgment become entangled. In particular, with regard to modality, this study argues that the Softmax function forcibly reframes AI judgments as possibility judgments. Furthermore, drawing on Kant's account of definition, this paper argues that no definitive criterion exists for verifying functional implementation. Moreover, fluent linguistic behavior may create the appearance of functional implementation even when important functions remain absent.
\end{abstract}


\section{Introduction}
\subsection{Why do we need a philosophy of AI?}
Philosophy has explored the concept of human intelligence for thousands of years, and the study of intelligence is a traditional philosophical topic. For example, logic attempts to explain the structure of knowledge, and epistemology seeks to understand how knowledge is obtained. In computer science, the powerful learning capabilities of deep learning \cite{lecun2015deep} have attracted attention. In particular, artificial neural networks (ANNs) \cite{lecun2015deep} are commonly used to mimic human intelligence. These are models that simply mimic the functions of neurons in the brain, making them mathematically tractable. From a mathematical perspective, these models are designed to perform desired intelligent functions by processing information through matrix multiplication involving many matrices. However, information regarding the type of judgment employed by AI remains insufficient. Indeed, the nature of the judgment made by AI is often considered a black box \cite{adadi2018peeking}. In the AI community, the term ``black box'' indicates that the reason behind ANNs' judgment is not presented in a manner comprehensible to humans. This phenomenon has been referred to as unexplainability in the AI community, which has led to the emergence of explainable AI as a field of research. Going further, it seems necessary to analyze this unexplainability more precisely using conceptual tools from philosophy, which have long engaged with the nature of intelligence.

\subsection{Is AI truly capable of making judgments?}
It may be argued that AI does not actually make judgments. It is not human. Nevertheless, the issue cannot be easily dismissed. Even if the notion of judgment cannot be applied to AI, we are still inevitably compelled to interpret its outputs through the lens of human reasoning. For instance, consider an AI model that outputs 1 when it detects a dog in an image. This numerical output is not understood as ``1'' itself; rather, humans interpret it as statements such as ``a dog is present in the image,'' ``there is a high probability that the image contains a dog,'' or ``a dog may be in the bottom left of the image.'' This reveals that we impose meaning from our own perspective. Such interpretations are independent of whether the machine is truly engaging in the act of judgment.

However, if machines do make judgments, a more serious problem arises: this leads to the question of whether explainable AI technology has fundamental limitations. We have striven to understand AI judgments within the confines of our human intellect. Can AI judgments possess additional characteristics beyond those of human judgments? Even if such judgments can exist, it appears that we lack the means to fully comprehend them. In that case, explainable AI technology is reduced to a tool that merely approximates machine judgments which fundamentally lie outside the bounds of human cognitive structures. Regardless of technological progress, true explainability may remain forever out of reach. If human cognition has inherent limitations and machines are able to surpass them, AI may come to occupy a unique position as an entirely new kind of epistemic agent.

I think that Turing also pursued his research \cite{turing1950computing} with a similar sense of problem awareness. He posed the question, ``Can machines think?'' but seemed to realize that it would be impossible to offer a universally valid and absolute proof to answer it. Thus, he shifted the direction of the question: even if the physical substance differs, what does it mean if, in terms of mental and functional capabilities, a machine and a human being are indistinguishable? He proceeded to develop his ideas from this new angle. This suggests that Turing departed from an ontological level of analysis. Ontology is a branch of philosophy concerned with what exists in the world, addressing the nature of being itself. For example, it asks questions such as “Does the number 1 exist?” or “Does color exist independently, or is it dependent on human perception?” This is a highly effective approach, as it enables one to develop discourse on AI without having to answer questions such as whether machines actually make judgments.

At the ontological level, we can pose questions about what the mind, mentality, or judgment itself is. One approach that seeks to address the nature of the mind at a structural level is functionalism \cite{levin2004functionalism}:

\begin{quote}
What makes something a thought, desire, pain (or any other type of mental state) depends not on its internal constitution, but solely on its function, or the role it plays, in the cognitive system of which it is a part.
\end{quote}

This claim posits that a system can be regarded as having a mind, even if it is not identical at the physical or chemical level, so long as it is identical in its functional aspects. Accordingly, it asserts that entities with biological structures different from those of humans, or even systems implemented on semiconductor hardware, may also be regarded as possessing a mind.

However, functionalism faces serious objections, particularly regarding whether functionality alone can account for genuine understanding. For instance, Searle's Chinese Room argument \cite{searle1980minds} demonstrates that the mind cannot be fully reduced to mere functionality. Simply put, just as one cannot be said to understand Chinese merely by receiving Chinese input and producing meaningful task outputs based on rules without knowing any Chinese, the mind cannot be reduced to a mere collection of functions. However, this is a study conducted at the ontological level, aiming to understand what the mind itself is. In other words, it is an analysis of the nature of the human mind. 

Whether functionalism itself has limitations poses no obstacle at all to research on AI. This is because we have not posed ontological questions about what the mind itself is. Instead, researchers need only focus on how to make what is human and what is machine functionally indistinguishable.

Escape from ontological questions is also evident in other cases. Observing the behavior of AI researchers today, they strive to artificially create entities that can be called ``intelligence'' and compete over how closely their creations resemble actual intelligence. When explaining AI systems that recognize and classify dogs, researchers often use terms related to intelligence such as judgment, pattern recognition, segmentation, and classification. However, when ordinary people look at such AI and say, ``There is no difference from humans,'' ``I see self-consciousness in it,'' researchers strongly refute such claims by insisting that AI is merely the result of mathematical computations. This contradiction is similar to the following: claiming ``Truth is relative to social, cultural, and historical contexts,'' yet demanding that this very claim be accepted as a universally valid truth, independent of any context.

Thus, this contradiction must be addressed first. AI researchers primarily focus on how to implement functionalities that resemble human intelligence. Therefore, even if their creations only approximate intelligence to a very small degree, they still call them AI. Every year, hundreds of papers are published in numerous journals to report new findings on neural networks in biology. Although we do not have a complete understanding of neural networks, we have made the definition of neural networks from our limited knowledge. Can such a definition be complete? Nevertheless, most people admit that ANNs have been created that resemble neural networks. In this sense, researchers are not discovering properties in an object, but rather assigning properties to it. In contrast, the general public observes the phenomena exhibited by AI and interprets them as the discovery of the properties of ``intelligence'' within it. For example, when seeing an AI that classifies images of dogs, people might say: ``this system has properties of human intelligence.'' 

This broader context ultimately compels not only researchers but also me to reconsider the nature of the questions we are asking. We cannot immediately conclude that AI judgment is, in the ontological sense, identical to genuine judgment itself. However, if the functional aspect of judgment has been implemented, it becomes an inevitable philosophical task to examine how closely this functional implementation resembles actual human judgment.

Because AI judgment does not need to be ontologically identical to actual judgment, there is the possibility that it could possess new properties. Functionally, AI judgment is not a complete replica of human judgment. However, partial similarity is sufficient for it to be recognized as an AI judgment. And, from this partial resemblance, new properties may arise. Thus, a distinct space for the philosophy of AI emerges, precisely because a fundamental difference between actual intelligence and AI is inherently embedded.

Thus, the philosophy of AI secures a distinct domain of inquiry, which will be elaborated in what follows. Investigating the limitations of AI --- determining which functions can never fully replicate human capabilities --- becomes a central problem within the philosophy of AI. At the same time, examining whether certain functions may transcend or overcome the limits of human intelligence also constitutes a core research topic within this newly emerging philosophical field. It is not necessary to engage with abstract notions such as whether machines can acquire ultimate knowledge beyond human intelligence. Intelligence consists of a variety of functions, including multiplication, an area in which AI has already demonstrated superiority over humans.

What I wish to state clearly is the following. Questions concerning whether machine intelligence can, in its essence or mode of being, become identical to human intelligence are beyond the scope of this paper. Likewise, claims that machines can never possess human intelligence because they lack a human body and do not possess intuitions of time and space, even if they implement functions that resemble human logical operations, are not the focus of the present study. For this reason, it is not necessary here to engage in investigations based on Kant's discussions of time, space, and the matter of sensibility presented in Kant's \textit{Critique of Pure Reason}.

Rather, the central issue addressed in this paper is how the outputs produced by machines are interpreted and explained through human cognition. The primary concern is not whether machine intelligence is ontologically identical to human intelligence, but how machine-generated results are interpreted, explained, and integrated within the cognitive framework through which human beings make sense of the world.

\section{Background}
This study mainly explores the nature of judgment, and the two figures who made the most significant contributions to this topic in the history of philosophy are Aristotle and Kant. To draw an analogy, their roles in the philosophy of judgment are comparable to Newton and Einstein in the history of physics. While this study is based on Kant’s framework, it should be noted that Kant’s system is not considered perfect, and ongoing research continues in the field of philosophy. One might wonder why a study on the nature of judgment would draw upon Kant’s framework, which was formulated centuries ago. Although both figures passed away long ago, subsequent research has largely focused on overcoming the frameworks established by these two giants. This situation is analogous to how, although the Transformer algorithm \cite{vaswani2017attention} was introduced nearly a decade ago, most mainstream follow-up research still focuses merely on modifying that original algorithm. It is worth noting that the Transformer architecture still fundamentally relies on the linear layer \cite{lecun2015deep}, a technique developed half a century ago. In this light, citing earlier philosophical frameworks such as Kant should not be hastily dismissed as outdated.

\subsection{Kant}
The English translation \cite{kant1855critique} is used in this work because Kant's \textit{Critique of Pure Reason} was not originally written in English. This paper investigates AI judgments, and for this purpose, it is necessary to clarify what a judgment is. Kant intricately maps and illustrates the features of judgment \citep[p. 34]{kant1855critique}. He presents Table \ref{table1}, which outlines the logical functions of judgment. Only through these forms can we engage in understanding and thought.

\begin{table}[h]
\begin{center}
\caption{Logical function of the understanding in judgments \citep[p. 34]{kant1855critique}.}
\label{table1}
\begin{tabular}{|p{0.15\columnwidth}|p{0.22\columnwidth}|p{0.21\columnwidth}|p{0.21\columnwidth}|}
\hline
\textbf{Quantity} & General & Special	& Individual \\
\hline
\textbf{Quality}&	Affirmative	& Negative	& Infinite\\
\hline
\textbf{Relation}& Categorical	& Hypothetical & Disjunctive\\
\hline
\textbf{Modality}& Problematic	& Assertoric & Apodictic\\
\hline
\end{tabular}
\end{center}
\end{table}

Each judgment corresponds to four categories --- one from each domain in Kant’s Table. In summary, every judgment necessarily falls into one of three categories in terms of quantity: whether the subject refers to a single entity, multiple entities, or all entities. From the perspective of quality, a proposition must either affirm a predicate, deny it, or negate a negation. In terms of relation, judgments express the relation between two concepts, the relation between two propositions, or the relation among multiple propositions. In terms of modality, there are judgments of possibility, actuality, and necessity. Table \ref{tableex} provides illustrative examples.

\begin{table}[h]
\centering
\caption{Kant's Types of Judgments with Examples}
\label{tableex}
\begin{tabular}{|p{0.3\columnwidth}|p{0.6\columnwidth}|}
\hline
\textbf{Type} & \textbf{Example} \\
\hline
\textbf{General} & All humans are mortal. \\
\textbf{Special} & Some animals are mammals. \\
\textbf{Individual} & Socrates is wise. \\
\hline
\textbf{Affirmative} & The apple is red. \\
\textbf{Negative} & The apple is not red. \\
\textbf{Infinite} & The apple is non-red (i.e., it belongs to the infinite set of things that are not red). \\
\hline
\textbf{Categorical} & The ball is round. \\
\textbf{Hypothetical} & If it rains, the ground gets wet. \\
\textbf{Disjunctive} & It is either day or night. \\
\hline
\textbf{Problematic} & It might be raining. \\
\textbf{Assertoric} & It is raining. \\
\textbf{Apodictic} & It must be raining, given the conditions. \\
\hline
\end{tabular}
\end{table}

Kant's categories could be applied to natural language processing \cite{chowdhary2020natural} in computer science. However, this has limitations due to the limitations of Kant's table, which is primarily designed for use with propositions having truth values. It seems to be applicable to the field of image captioning \cite{hossain2019comprehensive} with truth values. However, in general, orders and exclamations lack truth value in natural language. There are also instances in which propositions are hidden through irony. Consequently, natural language processing that is based on logic cannot possess the same capabilities as humans due to this inherent limitation.

From Table \ref{table1}, we can derive pure concepts of the understanding, which do not involve any empirical elements, and according to Kant, it is through these pure concepts that judgment is carried out. Table \ref{tableca} shows it.

\begin{table}[h]
\centering
\caption{Kant's Table of Categories \citep[p. 37]{kant1855critique}}
\label{tableca}
\begin{tabular}{|p{0.3\columnwidth}|p{0.6\columnwidth}|}
\hline
\textbf{Category Group} & \textbf{Categories} \\
\hline
\textbf{Quantity} & Unity, Multiplicity, Allness \\
\hline
\textbf{Quality} & Reality, Negation, Limitation \\
\hline
\textbf{Relation} & Substance and accident \\
 & Cause and effect \\
 & Community \\
\hline
\textbf{Modality} & Possibility — impossibility \\
 & Existence — nonexistence \\
 & Necessity — contingency \\
\hline
\end{tabular}
\end{table}

All the concepts presented in Table \ref{tableca} possess no empirical content whatsoever. For example, consider the concept of causality. Is causality something red? Is it large in size? We judge that one event is causally connected to another when we observe them repeatedly appearing in temporal succession. But, within the events themselves, no empirical material corresponding to ``cause'' can be found. Rather, it is only through such pure concepts that we are able to make causal judgments. Thus, these concepts belong to the form of thought required for experience.

For scientists, dealing with objects that lack empirical content might seem somewhat unfamiliar. However, in philosophy, there are fields that operate independently of experience, notably logic and metaphysics. Since no scientist denies the validity of logic, I will base my explanation on logic. If we open any standard textbook on logic, we find that propositions are typically represented by symbols such as A and B, and truth is inferred through logical symbols. What this means is that logic considers propositions only in terms of the aspects of thought that are entirely independent of content.

For instance, in the proposition A \(\lor \neg\) A (A or not A), it does not matter what specific content A represents. Whether A stands for ``the sun is composed of hydrogen and helium'' or ``all birds can fly'' is irrelevant. What is clear is that A \(\lor \neg A\) is always true, regardless of the actual content of A.

\section{Unexplainability of AI based on Kant's epistemology}
AI researchers naturally wonder why an AI system made a particular judgment. In doing so, they inquire not only into the surface outputs but into the underlying nature of the judgment itself. Once a system implements the functional structure of judgment, it is reasonable to assume that it also possesses, even if imperfectly, certain properties of judgment. Without such properties, the system's activity would cease to qualify as judgment in any meaningful sense. Thus, once we attribute the concept of ``judgment'' even partially to AI behavior, the system must, to varying degrees, embody the properties associated with judgment. As previously discussed, the functional realization of judgment may vary in completeness, ranging from weak to fully developed forms.

\subsection{Quantity, quality, and relation}
When AI makes a judgment, it produces numerical outputs. For instance, in image classification, let us say that if an image contains a face, the AI outputs 1; otherwise, it outputs 0. However, the determination of whether the judgment corresponds to the form of ``Universal," ``Particular," or ``Singular" in Table \ref{table1} is challenging. For instance, when an ANN learns to classify a human face, people may require additional information to distinguish among statements such as ``This image has a human face," ``Some of these types of images have a human face," and ``All these types of images have a human face." Human beings' expressions of their judgments are made in their exact form with natural language. However, AI outputs are represented as numbers. Therefore, when AI makes a specific judgment, we cannot classify its form in quantity in Table \ref{table1}. Furthermore, it is unclear which pure concepts of the understanding listed in Table \ref{tableca} should mediate the interpretation of AI outputs. In the following discussions, I will focus solely on the logical aspects without repeatedly mentioning the difficulty of mapping AI outputs to the pure concepts presented in Table \ref{tableca}.

In logic, only ``Affirmative" (A is B) and ``Negative" (A is not B) are deemed necessary. However, Kant distinguishes ``Infinite" from ``Negative" (A is (not B)) \citep[pp. 34-35]{kant1855critique}. It is not clear whether ANN outputs correspond to forms of ``Affirmative," ``Negative," or ``Infinite." For example, let us take again AI that is trained to output 1 when an image contains a human face and 0 when it does not. The judgments it makes during training with a value of 0 could be interpreted as ``This image has a nonhuman face" or ``This image does not have a human face."

Let me examine several cases in detail. While ``This image has a human face," ``This image does not have a human face," and ``This image has a nonhuman face" are all plausible interpretations, some are ruled out as a result of the characteristics of the task. First, we can imagine that the AI only uses the ``Affirmative" form of a judgment when it outputs 1 or 0. When it outputs 1, it states, ``This image has a human face." However, when it outputs 0, no predicate can be affirmed, except for the affirmation of the negative predicate itself. Hence, we conclude that ANN utilizes a combination of the ``Affirmative" and ``Infinite" forms of judgment. Second, we can imagine that the AI makes judgments that take the ``Negative" form. Here, an output of 0 would mean that ``This image does not have a human face," and an output of 1 would not produce any corresponding predicate, with the exception of the ``Affirmative" form. Hence, judgments are only possible involving a combination of ``Affirmative" and ``Negative" judgments. The use of the ``Infinite" form does not provide a solution. Additionally, with the exception of the two combinations of judgments determined above, a combination of the ``Infinite" and ``Negative" judgments would also seem impossible. If the ANN utilizes a combination of the ``Affirmative" and ``Infinite" forms of judgment, the question then arises: Is this type of learning acceptable? The ANN could be affirming that something is a nonhuman face but lacks further specification. From a logical point of view, the ``Infinite" form is identical to the ``Affirmative" form, and previous studies have found that ANNs perform their tasks effectively: They should output true values where we intend the output to be true and false where we intend the output to be false. However, due to the differences in content according to the quality of the judgments, ANNs may decide not to make judgments in the way that we intend.

Considering Kant's relation of judgment, this case presents even more difficult challenges than before. As summarized in Table \ref{table1}, ``Categorical" represents judgments on two concepts, ``Hypothetical" indicates the relation between the two judgments, and ``Disjunctive" signifies the relation between multiple judgments. For example, let's consider a model that classifies objects as either a car, a dog, or a cat. It's not always clear whether the model's judgment is based on a single proposition, two propositions, or multiple propositions. The judgment could simply be ``This image has a dog.'' It could also be ``If this image exhibits certain learned features, then it is judged to contain a dog.
In this case, the image satisfies those learned features. Therefore, the system judges that a dog is present.'' A hypothetical judgment is first established by linking certain learned features to the classification of a dog. Upon confirming that the conditions are satisfied, the system finalizes the judgment in accordance with the hypothetical structure. Also, it could be ``It is not a car and cat. I don't know what a dog is. But it has to be a dog anyway. Therefore, there must be a dog.'' In this last case, AI can make an accurate judgment without even possessing a full understanding of the concept of a dog. If this kind of learning occurs, it suggests that the learning process doesn't align with our intention. Considering that AI is typically trained to classify objects within specific predefined classes, this phenomenon is quite common. Even when employing various technologies to elucidate decisions by highlighting the dog's location in an image, it is uncertain whether AI comprehends the concept of a dog. This uncertainty arises because the AI might have made this determination based on distinguishing it from cats and cars.

\subsection{Modality}
According to Kant, the modality form has the following characteristics \citep[p. 35]{kant1855critique}.

\begin{quote}
The modality of judgments is a very special function of them, which has the distinguishing feature that it contributes nothing to the content of the judgment, (for apart from magnitude, quality, and proportion, there is nothing more that constituted the content of a judgment,) but only concerns the value of the copula in relation to thinking in general.
\end{quote}

An example of this is illustrated in Table \ref{table:modality_expressions}. Although the boiling point of water can vary depending on pressure, this factor is ignored here for simplicity. Moreover, whether the given proposition is actually true or false is not important in the present context.

\begin{table}[h]
\centering
\caption{Expression of the same content and a single event in different modalities}
\label{table:modality_expressions}
\begin{tabular}{|p{0.2\columnwidth}|p{0.7\columnwidth}|}
\hline
\textbf{Category} & \textbf{Expression} \\
\hline
\multicolumn{2}{|c|}{\textbf{(Fixed content with different modalities)}} \\
\hline
Problematic & Water can boil at 100 degrees Celsius. \\
\hline
Assertoric  & Water is boiling at 100 degrees Celsius. \\
\hline
Apodictic & Water must boil at 100 degrees Celsius.\\
\hline
\multicolumn{2}{|c|}{\textbf{(Single event with different modalities)}} \\
\hline
Problematic & The object can break upon impact. \\
\hline
Assertoric  & The object is currently falling due to gravity. \\
\hline
Apodictic & Upon analyzing the various intuitions given to me, it necessarily follows that an object falls due to gravity. \\
\hline
\end{tabular}
\end{table}

Kant’s theoretical framework emphasizes that while the content of a judgment remains fixed, the mode of its expression can vary. However, the focus of our discussion will be on the fact that a single phenomenon can be expressed through all three modalities. For example, consider a case where a computer solves an XOR logical problem. In human judgment, this is treated purely as a logical judgment, which is based on the modality of necessity. However, an AI system, which receives the input and output values as truth values, can approach the same problem either as a logical problem or as a classification task. It learns to produce the correct output depending on the input it receives. Thus, the outcome produced by AI can be: a logical consequence grounded in truth values (necessity), a judgment based on possibility by predicting probabilities through classification, or a classification judgment grounded in actuality, as it assigns the input to a specific class as a factual determination (e.g., ``It is A.'').

From this point, the probabilistic interpretation based on Softmax function \cite{sharma2017activation} begins to collapse. Softmax is a function that normalizes the outputs so that their sum becomes 1. Most AI researchers apply this function at the final stage, primarily for the purpose of probabilistic evaluation. Alternatively, as frequently used in attention mechanisms, it determines how multiple internal values are combined by assigning influence weights that sum to one. For example, in a simple image classification task distinguishing between dogs and cats, applying Softmax produces an output like 37\% for ``dog'' and 63\% for ``cat.'' Although this is a common example in computer vision, Softmax is almost always used in natural language processing (NLP) as well. Even though the core algorithms for NLP tasks constantly evolve, the fundamental approach remains the same: the problem is reduced to predicting the next word based on the previous context. For instance, given the phrase ``This morning I ate,'' the model predicts ``breakfast'' as the most probable next word. This prediction is expressed as probabilities over all possible outputs, which can number in the tens of thousands depending on the model. Of course, we can design many functions other than Softmax that normalize outputs to sum to one. However, Softmax remains the most widely used. What I want to point out is that this is not a story limited to Softmax. We often treat it as a kind of possibility judgment simply based on the condition that the sum of all possible cases equals 1. However, this reflects only a partial aspect of the full nature of possibility judgments, and it shows that we are not truly representing their complete characteristics. A possibility judgment does not necessarily have to be expressed in the form of a probability. For example, the statement ``We can draw a circle'' is a possibility judgment that is not quantified. In other words, we regard AI's possibility judgments as such based on partial functional similarity.

I would like to briefly step away from the main topic to explain why NLP is handled in that way. The idea that meaning is determined by the way language is utilized is well expressed in Wittgenstein's philosophy of language \cite{travis1989uses} and plays an important role in the philosophy of language as well. It is worth noting that natural language processing is also designed with the view that the meaning of a word is determined by the existence of surrounding words \cite{devlin2018bert, le2014distributed, mikolov2013efficient, vaswani2017attention}. It seems more convincing to say that they processed it in that way because the given data consisted of texts, rather than assuming they understood and implemented such a perspective from the philosophy of language.

Because it does not contribute to the content of the judgment, it allows us to explain it in the ``Problematical" form even if we do not know what judgment the machine makes. Therefore, even before the advent of explainable AI, it was possible to obtain judgments in the form of ``Problematical" only through the Softmax function. What we want to know through explainable AI may be the characteristics of the entire judgment shown in Table \ref{table1}, but another way is to artificially consider only the content of the judgment and intentionally fix the modality of the judgment. In other words, it is not that we will see what kind of judgment was made, but that we will treat it as a possibility judgment no matter what judgment was made. Softmax acts as an epistemological filter, compelling the interpretation of AI judgments in terms of possibility. As shown in Table \ref{table:modality_expressions}, even when the content of thought remains fixed, the modality of judgment may differ. If Softmax is further applied to the resulting judgment, it can transform even an assertoric statement, such as ``The object is currently falling,'' into a probabilistic form, e.g., ``The object might currently be falling with 70\% probability.'' Thus, even firm judgments can be reframed within a probability distribution. People might claim that we can transform some judgments into possibility judgments. For example, if something is said to be inevitably and logically true, the probability is 100\%. However, can ``actually exist," ``necessary," and ``possible" be the same? Treating all judgments as possibility judgments seems problematic.

Thus, we observe that unexplainability can exist in any of the four categories of judgment. This paper presents only a few examples, but unexplainability arises in infinitely many cases. Based on the examples provided here, it should be possible to generate many more new examples. Inspired by the uncertainty principle in quantum mechanics \cite{busch2007heisenberg}, we can call this AI's uncertainty. In this case, all forms of judgment overlap. The difficulty lies in determining which characteristics of judgment should be employed to interpret and explain such outputs within the structure of human cognition. AI judgments have advanced one step beyond mere indeterminacy — where they were simply unknown — by revealing that different forms of judgment are layered together.

In the field of explainable AI research, various technologies are being studied \cite{adadi2018peeking, selvaraju2017grad}. I believe that by developing technologies that provide insights into how AI judgments are made, we can achieve some level of explainability. Computer scientists have begun revealing which part of an image AI focuses on when making judgments. These tools assist us in inferring the subject and predicate. However, neither a subject nor a predicate is explicitly stated. For example, one representative technique involves highlighting which parts of an image led to the judgment that it is a dog. However, there are also areas such as image captioning, where a model generates captions based on a photo, and models that output both an answer and an explanation simultaneously \cite{sammani2022nlx}. These technologies explicitly generate natural language, producing both subjects and predicates. In doing so, AI begins to expand its mimicry beyond mere judgment (the content of thought) into the realm of reasoning (the act of thinking). The importance of such technology is particularly evident in domains like medical AI, where it is crucial to ensure that the rationale behind a model’s judgment aligns with the reasoning used by human physicians.

\section{Unexplainability related to functional implementation}
Johannes Hessen, in his textbook on philosophy \cite{hessen1947lehrbuch}, sharply distinguishes three essential components of thought. Thought consists of the act of thinking, the content of thought, and the object of thought. When we mentally consider something, this corresponds to the act of thinking. For example, in response to the proposition ``An elephant has four legs,'' one person may not be familiar with elephants and thus unable to visualize one internally, but can still picture an animal with four legs. Another person may mentally imagine an elephant and visualize its four legs. Yet another might picture a running elephant while reconstructing the proposition. In this way, even for a single proposition, the act of thinking occurs differently across individuals. However, the content of thought remains the same. What the proposition expresses is identically conveyed to multiple subjects; this is what we call the content of thought. Finally, the elephant and its legs --- as they exist --- appear in thought as the object of thought. Psychology addresses the act of thinking; logic is concerned with the content of thought; and ontology, along with the individual sciences, deals with the object of thought. The question of whether the content of thought corresponds to the object of thought always arises—and this is the domain of epistemology. For example, the question ``What is truth?'' belongs to this field. This summary encapsulates the information provided by Hessen and forms the conceptual foundation for the discussion that follows.

Through NLP technologies, we are now able to simultaneously address the act of thinking, the content of thought, and the object of thought. A model developed in recent research, powered by vast computational resources, has been trained on virtually unlimited amounts of human-generated text, thereby capturing how words are actually used. As a result, these systems now exhibit the ability to functionally replicate aspects of human natural language behavior with remarkable fluency. Such technologies are referred to as large language models (LLMs), with ChatGPT \cite{achiam2023gpt} being one of the most well-known examples. In the case of humans, it is impossible to conceive of thought content without the act of thinking—how can one be said to think without engaging in the process of thinking itself? Machines, however, are not bound to follow the human mode of thought. Nevertheless, because they learn from natural language, they indirectly acquire traces of human cognitive acts. For example, they can indirectly learn the act of thinking by learning how the flow of human thought is expressed through language. Natural language examples showing whether the content of thought matches its object allow for indirect learning about the object of thought. In keeping with the earlier perspective, we set aside ontological questions and focus solely on the functional aspects. That is, LLMs do not truly understand or possess the meanings of such words; instead, they learn how these words are used in relation to others, and are thus able to partially reproduce their functional behavior.

Even in the imitation of thought through NLP, we inevitably confront the unexplainability of AI, which arises from the intrinsic nature of language. According to Kant, a definition is always incomplete and can only express a concept to the extent that it is distinguishable \citep[pp. 212-213]{kant1855critique}. In short, one person may define ``gold'' by its color, weight, and chemical properties, while for another, the definition may omit weight and chemical characteristics. Since it is impossible to construct a fixed and rigorous definition with complete precision, the meanings of words cannot be permanently fixed. We do not possess a complete definition, and without it, we cannot determine the full function of a given word. Since it is impossible to construct a fixed and rigorous definition with complete precision, the meanings of words cannot be permanently fixed. Because concepts cannot be exhaustively defined, there exists no definitive criterion for determining whether a concept has been functionally implemented.

Even if one assumes that a concept can be fully defined, or that there exists a best possible definition that would be accepted by most humans, the fact that AI can freely and flawlessly use that concept in natural language across all cases in a manner indistinguishable from humans does not necessarily imply that the concept has been implemented. Through exposure to vast amounts of text, LLMs can freely use the concept of ``I.'' In this context, AI does not possess the same identity as a human being in terms of physical, chemical, or any other existential aspects; rather, it appears to have acquired only the ability to manipulate the concept of ``I'' to the extent that is functionally necessary in natural language. To be clear, this does not imply that the self in machines and the human self are identical as entities; rather, only certain functions associated with the self may be similar. The mere fact that both a car and human legs perform the function of locomotion does not entail that they are identical as entities. But can we determine whether the concept of the self has been functionally implemented through natural language? Even when we pose questions through natural language, we can observe that AI responds quite well, having been trained on vast amounts of data. However, such linguistic fluency alone does not provide sufficient grounds for concluding that the concept has been implemented. To illustrate this point, I will present a simple example based on \cite{seo2021neural}, suggesting that some self-related functions may extend beyond natural language.

\begin{figure}[h]
\centering
\includegraphics[width=\columnwidth]{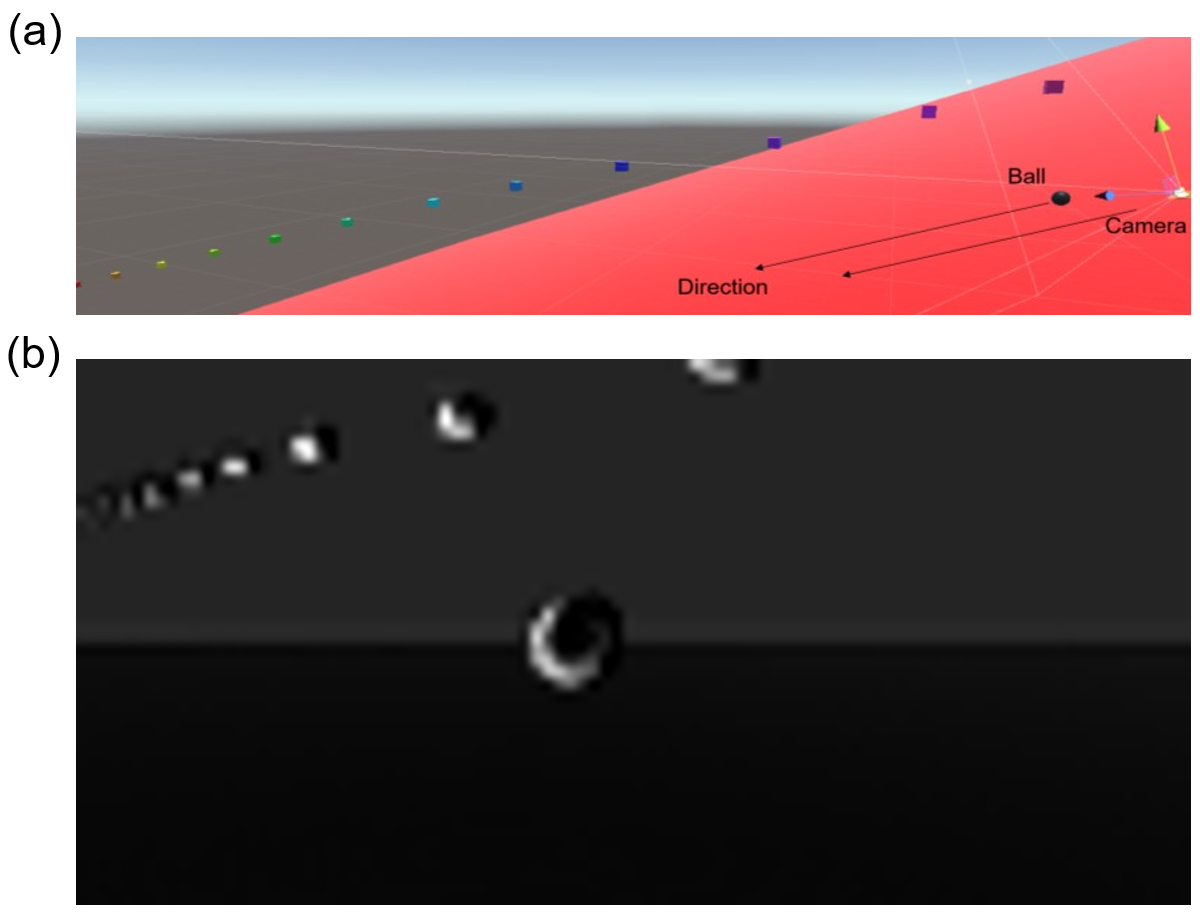} 
\caption{(a) A ball, a camera, and colorful boxes are fixed in the background. The task involves both the camera and the ball moving, and the goal is to match the ball’s velocity. The white areas indicate the regions where the AI is focusing its attention. Code and simulator are available at https://github.com/sjw007s/Unexplainability.}
\label{fig2}
\end{figure}

I conducted a brief additional experiment using a long-term recurrent convolutional network (LRCN) \cite{donahue2015long} and Grad-CAM \cite{selvaraju2017grad}. LRCN is an algorithm that learns from video input, where two-dimensional images are acquired sequentially over time and stacked. Grad-CAM is an algorithm that provides visual evidence for why an AI system made a particular decision. The target values were provided as a classification problem, aiming to predict whether the ball was faster or slower than the average speed.

The experiment reported in Figure \ref{fig2} may provide insight into the acquisition of self-related functions in humans and animals. Predator-prey situations offer an intuitive example in which distinguishing one's own movement from that of external objects becomes functionally important. In Figure \ref{fig2}, the ball and the camera may be interpreted as representing such a relationship, where successful behavior depends on tracking both self-generated and externally observed motion. This distinction may provide a functional basis for certain self-related functions.

Some studies \cite{bard2006self, gallup2002mirror, haikonen2007reflections} have suggested that the ability to recognize oneself in a mirror is associated with self-consciousness. However, mirrors are relatively uncommon in natural environments. From a functional perspective, interactions that require distinguishing self-generated motion from external motion may also contribute to the emergence of self-related functions. The present example is not intended to demonstrate self-consciousness itself, but rather to illustrate that certain self-related functions may extend beyond natural language.

In natural language, LLMs can use the concept of ``I'' with remarkable fluency and can even describe self-related functions in coherent sentences. However, successful performance in one domain does not by itself establish the implementation of the underlying concept. The realization of particular functions may obscure the absence of others. This illustrates the broader difficulty of verifying functional implementation, even when a system exhibits behavior that appears functionally indistinguishable from that of humans.

Even if functional implementation were achieved, it would establish only functional equivalence, not ontological identity. Functional similarity does not entail ontological identity. Thus, functional implementation alone provides no basis for concluding that the AI's ``I'' and the human ``I'' are identical.

\bibliography{aaai2027}


\end{document}